\documentclass[letterpaper]{article}
\usepackage{aaai2026}
\usepackage{times}
\usepackage{helvet}
\usepackage{courier}
\usepackage[hyphens]{url}
\usepackage{graphicx}
\usepackage{natbib}
\usepackage{caption}
\usepackage{amsmath}
\usepackage{amssymb}
\usepackage{booktabs}
\usepackage{multirow}
\usepackage{comment}
\nocopyright

\title{Calibrating WEAT Against Anisotropy: ZCA Whitening as a Geometric Pre-Processing Step for Embedding Association Tests}

\author{
    Seitaro Ono\textsuperscript{\rm 1},
    Senna Ross\textsuperscript{\rm 2},
    Jun Saiki\textsuperscript{\rm 1}
}
\affiliations{
    \textsuperscript{\rm 1}Graduate School of Human and Environmental Studies,
    Kyoto University, Kyoto, Japan\\
    \textsuperscript{\rm 2}Brock University, St. Catharines, Ontario, Canada\\
    ono@cv.jinkan.kyoto-u.ac.jp, sz23hr@brocku.ca, saiki.jun.8e@kyoto-u.ac.jp
}

\begin{document}
\maketitle

\begin{abstract}
We propose Zero-phase Component Analysis (ZCA) whitening 
as a geometric pre-processing step for the Word Embedding Association 
Test (WEAT). WEAT is a bias measurement method widely used in both 
computational social science and AI fairness research. It relies on 
cosine similarity as a measure of semantic association, which assumes 
that the embedding space is approximately isotropic. However, prior 
work has reported that many widely used language models do not 
satisfy this assumption, raising concerns about the reliability of bias measurements. ZCA whitening transforms the covariance of the embedding space into the identity matrix while minimizing perturbation to the original vectors. This transformation restores the isotropy condition on which WEAT relies. We evaluate our approach on ten standard WEAT test suites and seven 
models spanning three architectural families, yielding 70 model--task 
combinations. The results show that ZCA whitening substantially reduces 
the anisotropy of the embedding spaces across all models. 
Particularly for highly anisotropic models, we further observe 
improvements on standard semantic similarity benchmarks, indicating 
that the calibrated space better captures semantic associations. 
After calibration, over 30\% of WEAT results change significance 
status, and effect sizes shift in both directions depending on bias 
category. These shifts suggest that uncalibrated measurements may 
both overestimate and underestimate the associations encoded in the 
embedding space. These findings indicate that previously reported 
bias measurements in anisotropic embedding spaces should be 
interpreted with caution and may benefit from re-evaluation with 
calibrated methods. Our approach contributes to restoring the measurement foundation of WEAT across both computational social science and AI fairness research.
\end{abstract}

\begin{links}
    \link{Code}{https://github.com/seigit/zca-weat}
\end{links}

\noindent\textit{Accepted at the 9th AAAI/ACM Conference on AI, Ethics,
and Society (AIES 2026). This is an extended version including appendices;
the final published version will appear in the AAAI Digital Library.}

\begin{figure*}[t]
\centering
\includegraphics[width=0.6\textwidth]{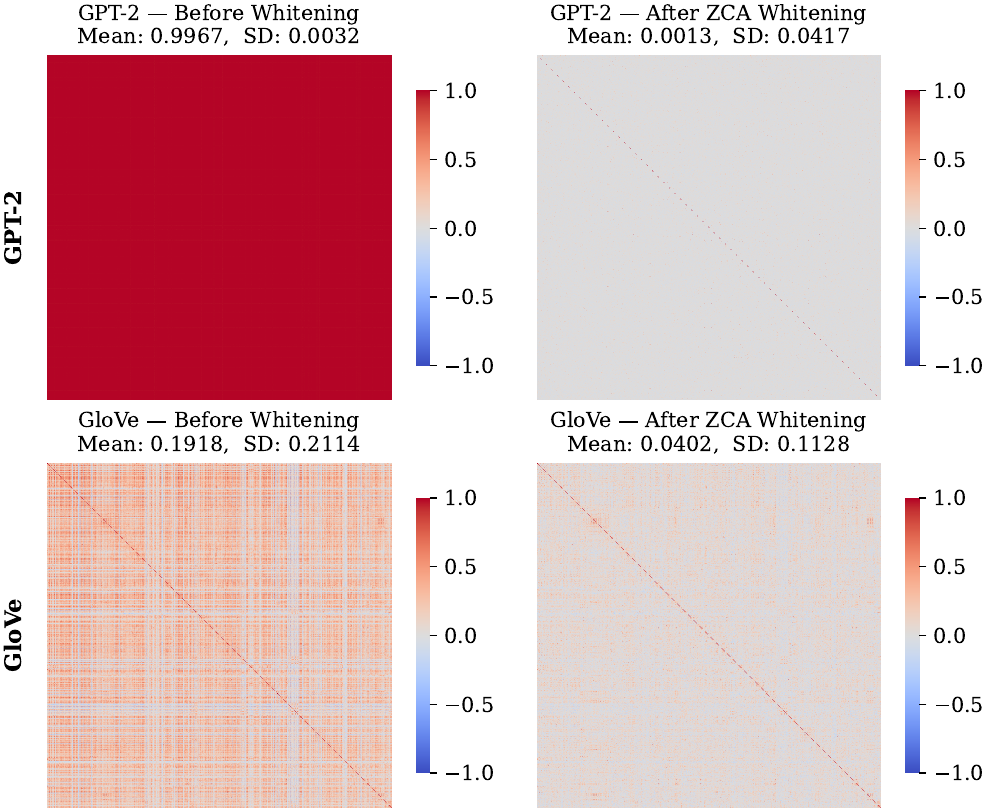}
\caption{Pairwise cosine similarity matrices for 2{,}000 randomly sampled embedding vectors from WikiText-103 before (left column) and after (right column) ZCA whitening. \textbf{Top:} GPT-2, whose extreme anisotropy (mean cosine similarity $= 0.997$) renders the matrix a uniform red field. \textbf{Bottom:} GloVe (mean $= 0.192$), which exhibits moderate anisotropy before whitening. After calibration, both models converge to near-zero mean similarity with meaningful pairwise variation restored.}
\label{fig:cosine_heatmap}
\end{figure*}

%==============================================================
\section{Introduction}
%==============================================================

Word embeddings \citep{mikolov2013distributed,pennington2014glove} encode statistical regularities of language, including social biases that can propagate to downstream applications \citep{bolukbasi2016man,caliskan2017semantics,garg2018word}.
Measuring these biases accurately is necessary for developing equitable NLP systems.
The Word Embedding Association Test \citep[WEAT;][]{caliskan2017semantics} transposes the logic of the Implicit Association Test \citep[IAT;][]{greenwald1998measuring} from human reaction-time studies to the geometry of word embedding spaces, measuring differential cosine-similarity associations between target concepts (e.g., European-American vs.\ African-American names) and attribute concepts (e.g., pleasant vs.\ unpleasant words).
The WEAT has attracted significant attention in both computational social science and AI fairness research, and has been applied in a large number of studies to measure racial, gender, age, and other biases across diverse models and modalities.

However, the validity of any measurement metric depends on the assumptions under which it operates.
WEAT assumes that cosine similarity faithfully captures semantic associations between words.
This assumption holds when the embedding space is approximately isotropic \citep{ethayarajh2019contextual,timkey2021all,rudman2022isoscore}.
In an isotropic space, vectors are distributed with roughly uniform variance in all directions. Under this condition, the angle between two vectors reflects their semantic relationship rather than a geometric artifact.

Recent work has shown that this isotropy assumption fails in many language models: static embeddings exhibit dominant mean directions \citep{mu2017all}, contextualized representations occupy narrow cones with near-unity pairwise similarities \citep{ethayarajh2019contextual}, and a small number of ``rogue dimensions'' dominate cosine similarity in Transformers \citep{timkey2021all}.

Despite this accumulating evidence, the consequences of anisotropy for bias measurement have received surprisingly little direct attention.
\citet{wolfe2024mleat} recently introduced ML-EAT, which provides a comprehensive multilevel evaluation framework and correctly identifies anisotropy as a threat to the validity of cosine-based bias tests.
However, the purpose of ML-EAT was to improve the interpretability of EAT measurements, and proposing a correction for anisotropy was outside its scope.
A calibration method that restores the geometric conditions under which cosine-based bias measurement becomes reliable remains a significant challenge.

In this paper, we introduce a calibration step that removes anisotropy from language model embedding spaces, restoring the isotropy assumption under which cosine-based bias measurement is valid.
Specifically, we propose ZCA (Zero-phase Component Analysis) whitening as a pre-processing calibration for embedding association tests.
ZCA whitening transforms the embedding space to have identity covariance while minimizing the perturbation to the original vectors, making it well suited for effectively reducing anisotropy.
Moreover, whitening has been shown to improve the quality of embeddings on standard semantic similarity benchmarks \citep{huang2021whiteningbert,su2021whitening}, suggesting that calibrated embedding spaces may better capture the semantic associations on which bias measurement depends.

We evaluate our approach across seven embedding models spanning three architectural families (static, contextualized, and contrastive unsupervised models) using the ten standard WEAT test suites from \citet{caliskan2017semantics}.
We make the following contributions:
\begin{enumerate}
\item We propose ZCA whitening as a novel pre-processing calibration method for embedding association tests and show that it substantially reduces anisotropy across all seven models in the embedding space.
\item We show that cosine similarity in the calibrated space more faithfully captures semantic associations on standard similarity benchmarks, supporting the validity of bias measurements in the calibrated space.
\item We systematically show that calibration changes WEAT outcomes in
both directions across 70 model--task combinations, reducing measured
associations in some configurations and increasing them in others.

\end{enumerate}

%\noindent Our research code will be made publicly available upon acceptance.

%==============================================================
\section{Related Work}
%==============================================================

\subsection{The Embedding Association Test}

\citet{caliskan2017semantics} introduced the Word Embedding Association Test (WEAT), a measurement of intrinsic bias in word embeddings drawing on the design of the IAT \citep{greenwald1998measuring}.
The WEAT quantifies the relative association of two target groups (such as European-American and African-American names) with two attribute groups (such as pleasant and unpleasant words).
Given target sets $X$ and $Y$ and attribute sets $A$ and $B$, the individual association for each target word $w$ is:
\begin{equation}
s(w, A, B) = \frac{1}{|A|}\sum_{a \in A} \cos(\vec{w}, \vec{a}) - \frac{1}{|B|}\sum_{b \in B} \cos(\vec{w}, \vec{b})
\label{eq:individual}
\end{equation}
The test statistic aggregates individual associations across the two target sets:
\begin{equation}
S(X, Y, A, B) = \sum_{x \in X} s(x, A, B) - \sum_{y \in Y} s(y, A, B)
\label{eq:teststat}
\end{equation}
The effect size $d$ normalizes this difference:
\begin{equation}
d = \frac{\operatorname{mean}_{x \in X}\, s(x, A, B) - \operatorname{mean}_{y \in Y}\, s(y, A, B)}{\operatorname{std}_{w \in X \cup Y}\, s(w, A, B)}
\label{eq:effectsize}
\end{equation}
This effect size is analogous to Cohen's $d$ \citep{cohen1992power}, where values of 0.2, 0.5, and 0.8 are conventionally interpreted as small, medium, and large effects.
Statistical significance is assessed via a one-sided permutation test that computes the $p$-value as:
\begin{equation}
p = \Pr_i\bigl[S(X_i, Y_i, A, B) > S(X, Y, A, B)\bigr]
\label{eq:permutation}
\end{equation}
where $\{(X_i, Y_i)\}$ denotes the set of all equally sized partitions of $X \cup Y$.

Since its introduction, the WEAT has been extended across 
modalities and architectures, in order to measure social biases in AI systems.
Text-based variants include SC-WEAT for single-target associations 
\citep{caliskan2017semantics}, SEAT for sentence-level embeddings 
\citep{may2019measuring}, a pretraining-objective-based adaptation 
\citep{kurita2019measuring}, and CEAT, which treats contextualization 
as a random effect at the word level \citep{guo2021detecting}.
The framework has also been extended beyond text to image encoders 
(iEAT; \citealt{steed2021image}), grounded vision-and-language models 
(Grounded-WEAT; \citealt{ross2021measuring}), CLIP 
\citep{wolfe2022american, 
wolfe2022hypodescent, radford2021learning}, speech models 
(SpEAT; \citealt{slaughter2023pre}), and text-to-video generators 
(VEAT; \citealt{sun2026veat}).
Recently, \citet{wolfe2024mleat} proposed ML-EAT, which 
provides multilevel interpretable bias measurements and includes 
an anisotropy diagnostic.
All of these variants rely on cosine similarity and are therefore 
subject to the same geometric artifacts that this paper addresses.
Their widespread adoption across NLP, computer vision, and speech 
processing underscores the importance of 
ensuring that their measurements are geometrically sound.

\subsection{Applications of EATs in Social Science}

Embedding association tests have been widely adopted in the social 
sciences as a tool for studying societal phenomena at scale.
A growing body of work uses EATs to probe contemporary cultural 
patterns in large corpora, including gender defaults and 
stereotypes in internet English \citep{caliskan2022gender, 
bailey2022based}, the relationship between gender bias and 
societal structure \citep{napp2023gender}, and temporal shifts 
in Wikipedia \citep{schmahl2020wikipedia}.
Cross-cultural and cross-linguistic studies have extended this 
reach to dozens of languages \citep{lewis2020gender, 
charlesworth2021gender, mukherjee2023global}.
Historical corpora have enabled the study of how these associations evolve 
over time, including a century of gender and ethnic stereotypes 
\citep{garg2018word}, 200 years of social-group stereotypes 
\citep{charlesworth2022historical}, intersectional stereotypes 
\citep{charlesworth2024extracting, borenstein2023measuring}, 
xenophobia in 19th--20th century travel literature 
\citep{sunsay2023historical}, and the expansion of the societal 
``moral circle'' \citep{leach2023word}.
Collectively, these applications demonstrate that EATs can serve not only as a measure of contemporary associations but also as a tool for tracing long-term societal change.

In applied domains, EATs have been used to measure biases in 
biomedical research \citep{rios2020quantifying}, clinical notes 
\citep{cobert2024measuring}, legal opinions 
\citep{matthews2022gender}, and court proceedings 
\citep{dutta2023gender}.
\citet{gray2025benchmarking} further proposed SD-WEAT, a variant 
handling multi-level attribute groups for healthcare bias 
benchmarking.
Recent work has also linked embedding-based measurements to human-level implicit associations, demonstrating meaningful correspondence between embedding biases and societal stereotypes \citep{charlesworth2024echoes, morehouse2023traces, manzini2019lipstick}.

These diverse applications underscore that the validity of EAT 
measurements is not a narrow technical concern but a matter of 
scientific integrity across multiple disciplines.
However, several studies have identified potential problems with 
methods that rely on cosine similarity between embedding vectors, 
suggesting that the geometric properties of the embedding space 
may affect the reliability of such measurements.

\subsection{Embedding Anisotropy and Its Implications for WEAT}

Anisotropy in word embeddings describes a geometric property in which vectors cluster in a narrow region of the embedding space \citep{cai2021isotropy}, causing cosine similarities between arbitrary word pairs to be systematically high.
\citet{ethayarajh2019contextual} demonstrated that contextualized representations from BERT, ELMo, and GPT-2 occupy a narrow cone with average cosine similarities approaching 1.0 in upper layers, while \citet{mu2017all} identified dominant principal components in word2vec and GloVe, and \citet{timkey2021all} showed that a few ``rogue dimensions'' dominate cosine computation in Transformers.
Further work has documented additional effects \citep{zhou2022problems,godey2024anisotropy,machina2024anisotropy,rajaee2022isotropy}.

For WEAT, such geometric distortion can compress or inflate the differential cosine associations that define its effect size, producing measurements that do not accurately reflect the underlying bias structure.
\citet{wolfe2024mleat} incorporated an anisotropy diagnostic into ML-EAT that flags unreliable models, but ML-EAT aims at interpretability rather than correction.
\citet{mu2017all} proposed removing top principal components, but this requires explicitly discarding information and selecting a model-dependent hyperparameter, limiting its practical applicability.
Thus, a calibration method that restores the reliability of cosine-based measurement without requiring such choices is desirable.

%==============================================================
\section{Approach}
%==============================================================

\subsection{How Anisotropy Distorts Cosine-Based Association Tests}

Cosine similarity between two vectors $\vec{u}, \vec{v} \in \mathbb{R}^d$ is defined as:
\begin{equation}
\cos(\vec{u}, \vec{v}) = \frac{\vec{u} \cdot \vec{v}}{\|\vec{u}\|\,\|\vec{v}\|}
\label{eq:cosine}
\end{equation}
For cosine similarity to serve as a reliable measure of semantic association, the embedding space should be approximately isotropic \citep{ethayarajh2019contextual,wolfe2024mleat}.
Specifically, the covariance matrix $\Sigma$ of the embedding vectors should be approximately proportional to the identity matrix~\citep{rudman2022isoscore}:
\begin{equation}
\Sigma \approx \sigma^2 I
\label{eq:isotropy}
\end{equation}

When $\Sigma$ departs from proportionality to $I$, cosine similarity may become systematically distorted.
Because cosine similarity is dominated by high-variance directions, vectors can appear more similar than they semantically are, while meaningful differences along low-variance directions are masked~\citep{mu2017all, timkey2021all}.

Since the WEAT computes differential cosine-similarity associations between target and attribute word sets (Equations~\ref{eq:individual}--\ref{eq:effectsize}), the effect size $d$ may be affected by anisotropy if the distortion is not uniform across the word sets involved in the test.
This suggests that anisotropy could both inflate and compress genuine bias signals, depending on the relationship between the word sets and the covariance structure of the embedding space.

To address this potential distortion, a calibration step that transforms the covariance matrix of the embedding space toward the identity matrix is needed, thereby restoring the geometric conditions under which cosine similarity can function as a reliable measure of semantic association.

\subsection{ZCA Whitening}

We adopt Zero-phase Component Analysis (ZCA) whitening as the pre-processing calibration step.

Given a set of embedding vectors $\{x_1, x_2, \ldots, x_n\} \subset \mathbb{R}^d$, we first compute the sample mean $\mu$ and the sample covariance matrix $\Sigma$:
\begin{equation}
\mu = \frac{1}{n}\sum_{i=1}^n x_i, \quad \Sigma = \frac{1}{n-1}\sum_{i=1}^n (x_i - \mu)(x_i - \mu)^\top
\label{eq:mean_cov}
\end{equation}
We then center each vector by subtracting the mean and apply the ZCA whitening matrix $W_{\text{ZCA}}$ to obtain the transformed vectors:
\begin{equation}
z_i = W_{\text{ZCA}}(x_i - \mu), \quad W_{\text{ZCA}} = \Sigma^{-1/2}
\label{eq:zca}
\end{equation}
where $\Sigma^{-1/2}$ denotes the symmetric matrix square root of $\Sigma^{-1}$.
To compute this matrix, we perform the eigendecomposition of the covariance matrix $\Sigma = U \Lambda U^\top$, where $U$ is an orthogonal matrix whose columns are the eigenvectors and $\Lambda$ is a diagonal matrix of the corresponding eigenvalues.
The whitening matrix then takes the explicit form:
\begin{equation}
W_{\text{ZCA}} = U \Lambda^{-1/2} U^\top
\label{eq:zca_explicit}
\end{equation}
This transformation rescales each eigenvector direction by the inverse square root of its eigenvalue: dimensions with disproportionately large variance are compressed, while dimensions with small variance are expanded, resulting in uniform variance across all directions.

The transformed vectors $\{z_i\}$ have identity covariance by construction:
\begin{equation}
\operatorname{Cov}(z) = W_{\text{ZCA}}\,\Sigma\, W_{\text{ZCA}}^\top = \Sigma^{-1/2}\, \Sigma\, \Sigma^{-1/2} = I,
\label{eq:identity_proof}
\end{equation}
where the final equality follows from the symmetry of $W_{\text{ZCA}} = \Sigma^{-1/2}$ (see Appendix~A for a full derivation).
Since $\operatorname{Cov}(z) = I$, the calibrated embedding space satisfies the isotropy condition (Equation~\ref{eq:isotropy} with $\sigma^2 = 1$) under which cosine similarity can function as a reliable measure of semantic association.

Among the family of whitening transformations that produce 
identity covariance, ZCA uniquely minimizes the expected squared 
distance between original and transformed vectors 
\citep{kessy2018optimal}, making it particularly suitable for 
bias measurement calibration where preserving the original 
semantic structure is essential.
Moreover, whitening has been shown to substantially improve 
scores on standard semantic textual similarity benchmarks 
\citep{huang2021whiteningbert, su2021whitening}, suggesting 
that the calibrated embedding space more faithfully captures 
semantic associations between words and may therefore provide 
a more reliable basis for extracting social associations that 
were obscured by anisotropy in the original space.

\subsection{Calibration Pipeline}

Our full measurement pipeline operates in three stages.

In the first stage, we estimate the whitening matrix from a large reference sample of embedding vectors.
We adopt WikiText-103 \citep{merity2017pointer} as the reference corpus for collecting these samples.
From this corpus, we draw 100,000 samples and feed them into the model under evaluation to obtain the corresponding embedding vectors.
Using the resulting embeddings, we compute the sample mean $\mu$ and the covariance matrix $\Sigma$ (Equation~\ref{eq:mean_cov}), from which we estimate the ZCA whitening matrix $W_{\text{ZCA}}$ (Equation~\ref{eq:zca_explicit}).
This estimation is performed once per model and the resulting whitening statistics are reused across all test suites.

In the second stage, we apply the estimated whitening transformation to the specific word vectors involved in the WEAT test suites.
Each word vector is centered by subtracting the reference mean $\mu$ and then transformed by $W_{\text{ZCA}}$ to obtain the calibrated vector (Equation~\ref{eq:zca}).

In the third stage, we compute WEAT scores (effect size $d$ and permutation $p$-value) using the calibrated vectors.

This pipeline is modular: it can be applied to any embedding model and any cosine-similarity-based bias metric without modifying the metric itself.

%==============================================================
\section{Experimental Setup}
%==============================================================

\begin{table*}[t!]
\centering
\caption{The ten standard WEAT test suites \citep{caliskan2017semantics}. Each suite specifies two target sets ($X$, $Y$) and two attribute sets ($A$, $B$). $|{\cdot}|$ denotes the number of stimulus words in each set, following the lists provided in \citet{caliskan2017semantics}. EA and AA denote European-American and African-American, respectively.}
\label{tab:weat_suites}
\small
\begin{tabular}{@{}clllllcccc@{}}
\toprule
\textbf{ID} & \textbf{Bias Type} & \textbf{Target $X$} & \textbf{Target $Y$} & \textbf{Attribute $A$} & \textbf{Attribute $B$} & $|X|$ & $|Y|$ & $|A|$ & $|B|$ \\
\midrule
W1  & Valence    & Flowers        & Insects        & Pleasant   & Unpleasant   & 25 & 25 & 25 & 25 \\
W2  & Valence    & Instruments    & Weapons        & Pleasant   & Unpleasant   & 25 & 25 & 25 & 25 \\
W3  & Race       & EA names       & AA names       & Pleasant   & Unpleasant   & 32 & 32 & 25 & 25 \\
W4  & Race       & EA names       & AA names       & Pleasant   & Unpleasant   & 16  & 16  & 25 & 25 \\
W5  & Race       & EA names       & AA names       & Pleasant   & Unpleasant   & 16  & 16  & 8  & 8  \\
W6  & Gender     & Male names     & Female names   & Career     & Family       & 8  & 8  & 8  & 8  \\
W7  & Gender     & Math           & Arts           & Male terms & Female terms & 8  & 8  & 8  & 8  \\
W8  & Gender     & Science        & Arts           & Male terms & Female terms & 8  & 8  & 8  & 8  \\
W9  & Health     & Mental disease & Physical disease & Temporary & Permanent   & 6  & 6  & 7  & 7  \\
W10 & Age        & Young names    & Old names      & Pleasant   & Unpleasant   & 8  & 8  & 8  & 8  \\
\bottomrule
\end{tabular}
\end{table*}

\subsection{Models}
We evaluate seven embedding models spanning three architectural families.

\paragraph{Static embeddings.}
GloVe \citep{pennington2014glove} trained on Wikipedia and Gigaword (6B tokens, 100 dimensions), and word2vec \citep{mikolov2013distributed} trained on Google News (approximately 100B tokens, 300 dimensions).
We use the 6B-token, 100-dimensional GloVe model rather than the 
840B-token, 300-dimensional variant used in \citet{caliskan2017semantics}.
Our purpose is not to replicate their specific results but to evaluate the impact of anisotropy calibration, and the 6B model exhibits sufficient anisotropy (Table~\ref{tab:anisotropy}) to serve this purpose.

\paragraph{Contextualized embeddings.}
BERT-base-cased \citep{devlin2019bert}, RoBERTa-base \citep{liu2019roberta}, and GPT-2 \citep{radford2019language}, each with 12 Transformer layers and 768-dimensional hidden representations.
For all Transformer models, we use representations from the final layer, consistent with prior work applying EATs to contextualized models \citep{may2019measuring,guo2021detecting,wolfe2024mleat}.
Word-level embeddings are obtained following the Aggregated procedure of \citet{bommasani2020interpreting}: for each target or attribute word, we sample up to 20 sentences containing that word from a preprocessed WikiText-103 corpus.
Preprocessing retains sentences of 7--75 tokens and excludes section headers, yielding up to 200{,}000 indexed sentences with a fixed random seed for reproducibility.
Within each sentence, subword tokens of the target word are mean-pooled into a single contextualized vector, and the resulting vectors across the 20 sentences are then averaged to produce one word-level representation.
\citet{bommasani2020interpreting} reported that bias evaluation results were fairly stable across $n_i \in \{20, 50, 100\}$ contexts per word. We therefore adopted $n_i = 20$ to maintain consistency with their protocol and to minimize computational overhead.
We applied the canonical WEAT to these word-level representations rather than sentence-level variants such as SEAT \citep{may2019measuring} or distributional variants such as CEAT \citep{guo2021detecting}. This choice allows us to use a unified measurement procedure across all seven models, including static embeddings for which SEAT and CEAT are not applicable.

\paragraph{Unsupervised contrastive embeddings.}
We adopt two unsupervised contrastive models from the SimCSE framework \citep{gao2021simcse}: unsup-SimCSE-BERT-base-uncased (unsup-BERT) and unsup-SimCSE-RoBERTa-base (unsup-RoBERTa).
Both models share the same Transformer architecture as their supervised counterparts (12 layers, 768-dimensional hidden representations), and word-level embeddings are obtained using the same aggregation procedure of \citet{bommasani2020interpreting} described above.
These models differ from the standard contextualized models in two 
key aspects: training signal (unsupervised vs.\ supervised) and 
objective function (contrastive vs.\ language modeling). We include 
them in our analysis to examine how these factors influence the 
degree of anisotropy and the patterns of bias detected by WEAT.

\paragraph{Whitening statistics.}
The whitening matrix $W_{\text{ZCA}}$ is estimated from WikiText-103 \citep{merity2017pointer}.
For static models, we sample up to 100,000 vocabulary items.
For contextualized and unsupervised contrastive models, we sample up to 100,000 sentences.
These contextualized models are pretrained 
on sentence-level inputs. Feeding isolated words may therefore 
produce out-of-distribution representations \citep{bommasani2020interpreting}. 
To avoid this issue, we use sentence-level inputs to estimate the 
covariance matrix.
To ensure numerical stability when computing the whitening matrix (Equation~\ref{eq:zca_explicit}), we regularize the eigenvalues by replacing $\Lambda$ with $\Lambda + \varepsilon I$ before inversion, yielding:
\begin{equation}
W_{\text{ZCA}} = U (\Lambda + \varepsilon I)^{-1/2} U^\top
\label{eq:zca_regularized}
\end{equation}
where $\varepsilon = 10^{-3}$.
This prevents near-zero eigenvalues from producing excessively large entries in the whitening matrix.

\subsection{WEAT Test Suites}

We use the ten standard WEAT test suites from \citet{caliskan2017semantics} (Table~\ref{tab:weat_suites}).
These test suites correspond to well-established IAT findings in social psychology and have been used in numerous studies to benchmark bias in NLP models \citep{kurita2019measuring,may2019measuring,guo2021detecting,wolfe2024mleat}.

\begin{figure*}[t]
\centering
\includegraphics[width=0.9\textwidth]{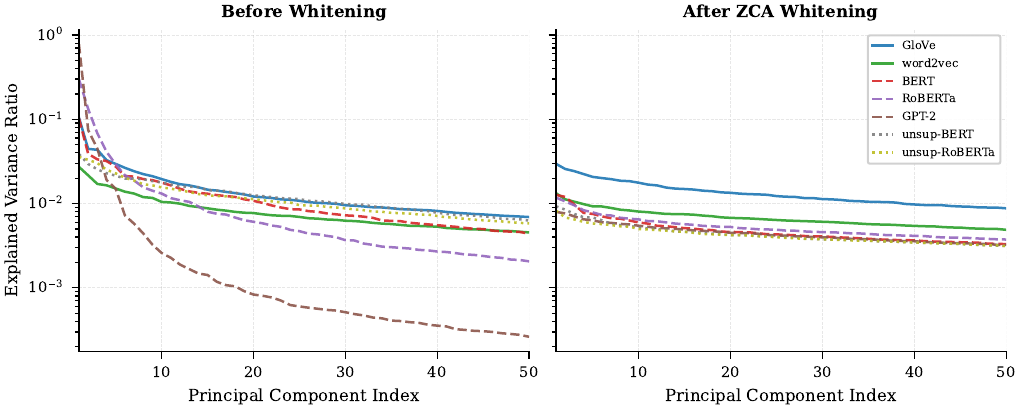}
\caption{Explained variance ratio of the top 50 principal components (log scale) before (left) and after (right) ZCA whitening for all seven models. \textbf{Left:} Contextualized models (GPT-2, RoBERTa, BERT) show extreme variance concentration in the first few components, while static models (GloVe, word2vec) and contrastive models (unsup-BERT, unsup-RoBERTa) exhibit flatter distributions. \textbf{Right:} After whitening, all models converge to a near-uniform eigenvalue distribution.}
\label{fig:eigenvalue}
\end{figure*}

\subsection{Evaluation Protocol}

\paragraph{Experiment 1: Anisotropy measurement.}
We quantify anisotropy by computing the mean and standard deviation of pairwise cosine similarities among 2,000 randomly sampled embedding vectors from WikiText-103, both before and after ZCA whitening.
An isotropic space yields a mean pairwise cosine similarity near zero.
In addition, we qualitatively compare the embedding spaces before and after whitening by examining the explained variance ratio of the covariance matrices, which illustrates how variance is distributed across dimensions. We also visualize pairwise cosine similarity matrices over the sampled vectors.

\paragraph{Experiment 2: Impact of anisotropy on semantic structure.}
We evaluate how anisotropy affects the quality of cosine-similarity-based semantic measurements by computing Spearman rank correlations ($\rho$) on WordSim-353 \citep{finkelstein2002placing}, SimLex-999 \citep{hill2015simlex}, and STS-B \citep{cer2017semeval} before and after whitening.
Each benchmark provides word or sentence pairs with human-assigned similarity scores. A higher Spearman correlation indicates that cosine similarity better captures the semantic associations in these benchmarks.
Statistical significance is assessed via paired bootstrap resampling (10,000 iterations).
If calibration improves these correlations, it suggests that cosine similarity in the calibrated space more faithfully captures semantic associations, supporting the validity of bias measurements conducted in that space.

\paragraph{Experiment 3: Bias measurement with WEAT.}
For each model--task combination, we compute the WEAT effect size $d$ and permutation $p$-value (100,000 permutations) on both raw and whitened embeddings.
Results are classified based on the change in statistical significance at $p < 0.05$:
\textbf{Stable}, significance unchanged;
\textbf{Disappearing}, significant before whitening but non-significant after;
\textbf{Emerging}, non-significant before but significant after whitening.
We then analyze, for each bias category, how these classifications distribute across models and identify the systematic patterns of distortion induced by anisotropy.

%========================================
\section{Results}
%==============================================================

\begin{table}[t]
\centering
\caption{Anisotropy measured by mean pairwise cosine similarity 
($\pm$ standard deviation, SD) among 2,000 randomly sampled vectors 
before (Raw) and after (White) ZCA whitening.}
\label{tab:anisotropy}
\small
\setlength{\tabcolsep}{4pt}
\begin{tabular}{@{}lcccc@{}}
\toprule
\textbf{Model} & \textbf{Raw Mean} & \textbf{Raw SD} & \textbf{Wh.\ Mean} & \textbf{Wh.\ SD} \\
\midrule
GloVe        & 0.192 & 0.211 & 0.040 & 0.113 \\
word2vec     & 0.059 & 0.074 & 0.013 & 0.067 \\
BERT         & 0.821 & 0.068 & 0.002 & 0.043 \\
RoBERTa      & 0.954 & 0.024 & 0.001 & 0.048 \\
GPT-2        & 0.997 & 0.003 & 0.001 & 0.042 \\
unsup-BERT   & 0.386 & 0.094 & 0.001 & 0.042 \\
unsup-RoBERTa & 0.475 & 0.090 & 0.001 & 0.041 \\
\bottomrule
\end{tabular}
\end{table}

\subsection{ZCA Whitening Reduces Anisotropy}

Table~\ref{tab:anisotropy} presents the mean pairwise cosine similarity before and after ZCA whitening for all seven models.
Static embeddings exhibit mild anisotropy (GloVe: 0.192; word2vec: 0.059), while contextualized models display severe anisotropy, with GPT-2 reaching an extreme value of 0.997 (SD = 0.003) in which virtually all vectors point in nearly the same direction.
This pattern is consistent with prior reports of cosine similarities concentrating near 1.0 between arbitrary word pairs in GPT-2 Base \citep{timkey2021all, wolfe2024mleat}.
The unsupervised contrastive models show intermediate anisotropy (unsup-BERT: 0.386; unsup-RoBERTa: 0.475), consistent with the uniformity-promoting nature of contrastive learning \citep{gao2021simcse} but still substantial enough to distort cosine-based measurements.
After ZCA whitening, all models converge to near-zero mean pairwise cosine similarity (0.001--0.040), confirming that calibration reduces anisotropy regardless of the initial degree of distortion. This reduction is also visible in the pairwise cosine similarity matrices (Figure~\ref{fig:cosine_heatmap}). GPT-2's matrix appears as a nearly uniform red field before whitening, while GloVe exhibits moderate anisotropy. After calibration, both center near zero with restored pairwise variation. The same qualitative pattern holds for the other five models (see Appendix~B).

Figure~\ref{fig:eigenvalue} provides further confirmation through the explained variance ratio of each model's embedding covariance matrix.
Before whitening, the contextualized models show extreme variance concentration in the first few principal components, whereas the static and contrastive models exhibit flatter but still non-uniform distributions.
After whitening, the explained variance ratio flattens substantially across all models, indicating that dominant directional components have been removed and variance is distributed uniformly across dimensions.

\subsection{Calibrated Embeddings Better Capture Semantic Associations}

Table~\ref{tab:semantic} reports Spearman rank correlations on three semantic similarity benchmarks before and after ZCA whitening.
For models with high anisotropy (GPT-2, RoBERTa, and BERT), whitening produces substantial improvements on word-level benchmarks (WordSim-353 and SimLex-999).
The most dramatic gains appear for GPT-2, where Spearman $\rho$ on WordSim-353 increases from 0.263 to 0.620 ($\Delta = +0.358$, $p < 0.001$) and on SimLex-999 from 0.097 to 0.411 ($\Delta = +0.314$, $p < 0.001$), suggesting that the original embedding space was so anisotropic that cosine similarity was largely non-functional and whitening restored much of its discriminative capacity.
This pattern is consistent with \citet{timkey2021all}, who showed that correcting for rogue dimensions substantially improves cosine similarity as a measure of semantic association.
RoBERTa and BERT show similar word-level gains ($\Delta = +0.054$ to $+0.191$), while their STS-B scores are either unchanged (RoBERTa) or modestly decreased (BERT: $\Delta = -0.055$).

For word2vec, which has the lowest initial anisotropy (mean cosine 0.059), the results are mixed and small in magnitude ($|\Delta| \leq 0.027$), suggesting that whitening neither substantially improves nor degrades an already near-isotropic space.
The unsupervised contrastive models show essentially unchanged word-level scores but small, significant decreases on STS-B ($\Delta = -0.055$ for unsup-BERT; $\Delta = -0.034$ for unsup-RoBERTa), possibly reflecting a domain mismatch between the reference corpus (WikiText-103) and the data on which these models were originally optimized.

Taken together, these results indicate that ZCA whitening preserves the semantic structure of the embedding space. In many cases, it also enables cosine similarity to more faithfully capture semantic associations. This improvement is particularly pronounced for highly anisotropic models, where calibration is most needed.
The modest STS-B decreases warrant acknowledgment but are small relative to the word-level gains.
Because WEAT measures bias through cosine similarity, these findings support the interpretation that WEAT measurements in the calibrated space are more reliable indicators of genuine semantic associations, particularly for models with high initial anisotropy.

\begin{table}[t]
\centering
\caption{Spearman $\rho$ on semantic similarity benchmarks before (Raw) and after (White) ZCA whitening. $\Delta$ denotes the change. Significance: $^{***}p < 0.001$, $^{**}p < 0.01$, $^{*}p < 0.05$, n.s.\ = not significant (paired bootstrap, 10k iterations).}
\label{tab:semantic}
\small
\begin{tabular}{@{}llrrrl@{}}
\toprule
\textbf{Model} & \textbf{Bench.} & \textbf{Raw} & \textbf{White} & \textbf{$\Delta$} & \textbf{Sig.} \\
\midrule
\multirow{3}{*}{GloVe}
 & WS-353 & .477 & .586 & +.109 & $^{***}$ \\
 & SL-999 & .297 & .380 & +.083 & $^{***}$ \\
 & STS-B  & .560 & .599 & +.039 & $^{***}$ \\
\midrule
\multirow{3}{*}{word2vec}
 & WS-353 & .686 & .661 & $-$.025 & $^{*}$ \\
 & SL-999 & .441 & .469 & +.027 & $^{***}$ \\
 & STS-B  & .707 & .684 & $-$.023 & $^{***}$ \\
\midrule
\multirow{3}{*}{BERT}
 & WS-353 & .560 & .658 & +.098 & $^{***}$ \\
 & SL-999 & .412 & .465 & +.054 & $^{***}$ \\
 & STS-B  & .633 & .578 & $-$.055 & $^{***}$ \\
\midrule
\multirow{3}{*}{RoBERTa}
 & WS-353 & .446 & .559 & +.113 & $^{***}$ \\
 & SL-999 & .266 & .457 & +.191 & $^{***}$ \\
 & STS-B  & .650 & .648 & $-$.002 & n.s. \\
\midrule
\multirow{3}{*}{GPT-2}
 & WS-353 & .263 & .620 & +.358 & $^{***}$ \\
 & SL-999 & .097 & .411 & +.314 & $^{***}$ \\
 & STS-B  & .428 & .596 & +.168 & $^{***}$ \\
\midrule
\multirow{3}{*}{\shortstack[l]{unsup-\\BERT}}
 & WS-353 & .718 & .734 & +.016 & n.s. \\
 & SL-999 & .536 & .545 & +.009 & n.s. \\
 & STS-B  & .840 & .786 & $-$.055 & $^{***}$ \\
\midrule
\multirow{3}{*}{\shortstack[l]{unsup-\\RoBERTa}}
 & WS-353 & .536 & .532 & $-$.004 & n.s. \\
 & SL-999 & .436 & .407 & $-$.029 & n.s. \\
 & STS-B  & .845 & .811 & $-$.034 & $^{***}$ \\
\bottomrule
\end{tabular}
\end{table}

\subsection{Anisotropy Distorts WEAT in Both Directions}

\begin{figure*}[t!]
\centering
\includegraphics[width=0.98\textwidth]{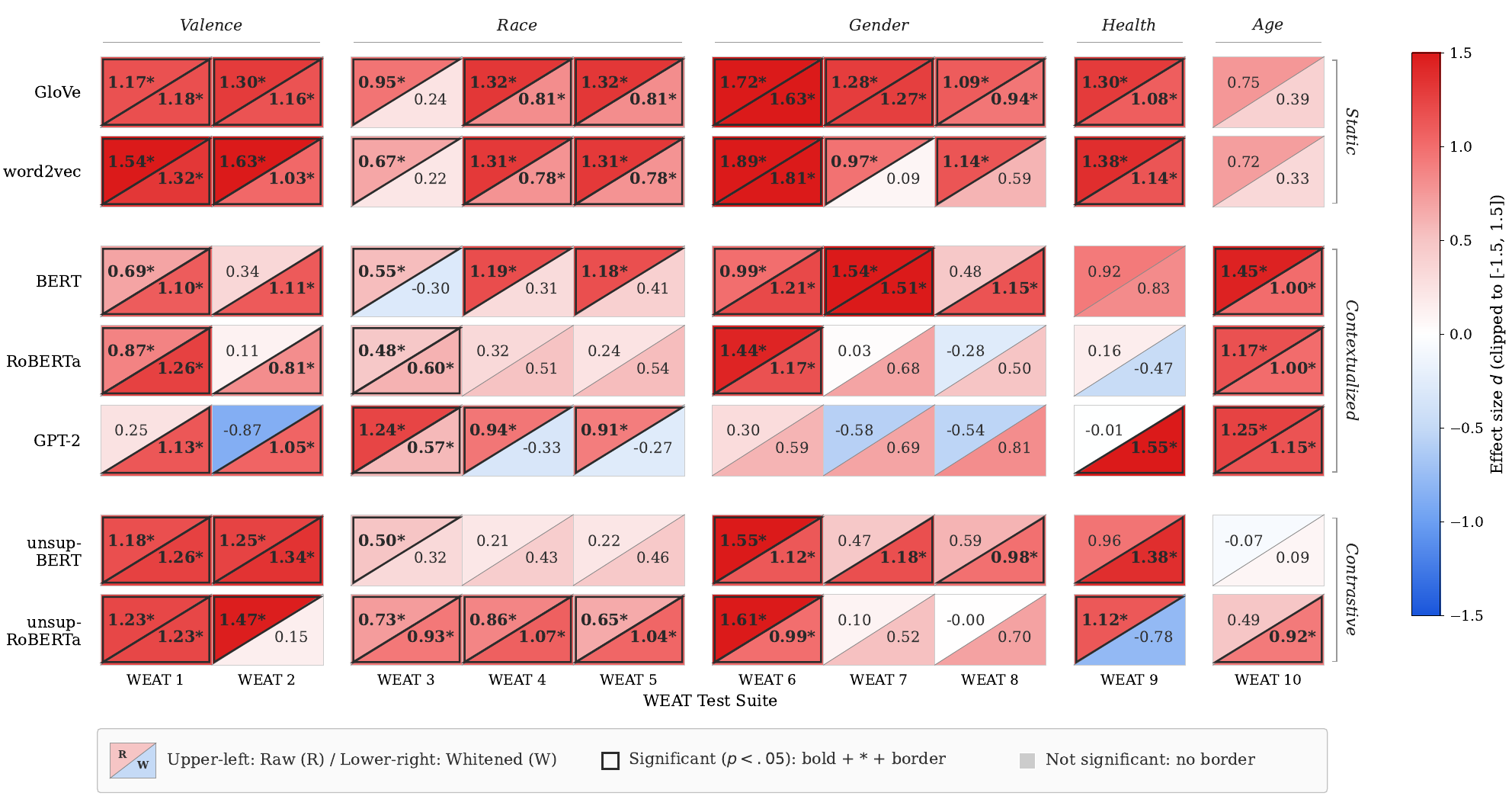}
\caption{WEAT effect sizes ($d$) before (upper-left triangle) and after 
(lower-right triangle) ZCA whitening across seven models and ten test suites. 
Color intensity encodes effect size magnitude (clipped to $[-1.5, 1.5]$). 
Bold values with asterisks and bordered cells indicate statistical 
significance ($p < 0.05$); regular-weight values without borders indicate 
non-significant results. Models are grouped by architectural family: 
static (top), contextualized (middle), and unsupervised contrastive (bottom). 
Test suites are grouped by bias type: Valence (W1--W2), Race (W3--W5), 
Gender (W6--W8), Health (W9), and Age (W10).}
\label{fig:weat_heatmap}
\end{figure*}

Figure~\ref{fig:weat_heatmap} visualizes WEAT effect sizes before and after ZCA whitening across all 70 model--task combinations, and Figure~\ref{fig:weat_scatter} plots whitened versus raw effect sizes grouped by bias type.
Full numerical results, including effect sizes, $p$-values, and significance change classifications for all combinations, are provided in Appendix~C.
Across the 70 combinations, we observe 48 Stable cases (68.6\%), 12 Disappearing cases (17.1\%), and 10 Emerging cases (14.3\%). Effect size shifts also occur within Stable cases. Among the Stable cases that remain significant in both spaces, 7 exhibit
$|\Delta d| > 0.5$, including GPT-2 on W3 ($d = 1.24 \to 0.57$) and unsup-RoBERTa on W6 ($d = 1.61 \to 0.99$). In total, 29 of 70 combinations (41.4\%) show either a significance change or a substantial
effect-size shift ($|\Delta d| > 0.5$) after calibration, suggesting that
anisotropy is a practical source of measurement error.

\subsubsection{Valence (W1--W2).}

Most valence measurements cluster near or above the diagonal in Figure~\ref{fig:weat_scatter}(a), indicating that effect sizes are preserved or increase after whitening.
Static models retain significance throughout, though word2vec shows compression of inflated effects (W2: $d = 1.63 \to 1.03$). The contextualized models show a markedly different pattern: four of six contextualized model--task combinations exhibit Emerging bias (BERT W2, RoBERTa W2, GPT-2 W1, GPT-2 W2). The most striking case is GPT-2 on W2 ($d = -0.87 \to d = 1.05$). BERT and RoBERTa on W2, and GPT-2 on W1, follow the same pattern (see Appendix~C).
Because valence associations such as flowers--pleasant and weapons--unpleasant are among the most robust findings in the IAT literature \citep{greenwald1998measuring}, their absence in uncalibrated contextualized embeddings likely reflects geometric distortion rather than a genuine lack of association.
The contrastive models are mostly Stable, with one Disappearing case (unsup-RoBERTa W2: $d = 1.47 \to 0.15$).

\subsubsection{Race (W3--W5).}

Race measurements are predominantly located below the diagonal in Figure~\ref{fig:weat_scatter}(b), indicating systematic decreases in effect size after whitening.
This category concentrates the bulk of overestimation: 8 of 12 Disappearing cases (66.7\%) occur in W3--W5.
Both static models show Disappearing bias on W3 (e.g., GloVe: $d = 0.95 \to 0.24$), with reduced but still significant effects on W4 and W5.
Among the contextualized models, BERT exhibits Disappearing bias across all three race tests, and GPT-2 on W4 and W5 (full transitions in Appendix~C); RoBERTa is the exception, maintaining stable significance throughout.
The contrastive models show only one Disappearing case within the race tests (unsup-BERT W3), and unsup-RoBERTa even shows slightly increased race effects after whitening (W4: $d = 0.86 \to 1.07$).

\subsubsection{Gender (W6--W8).}

\begin{figure*}[t!]
\centering
\includegraphics[width=\textwidth]{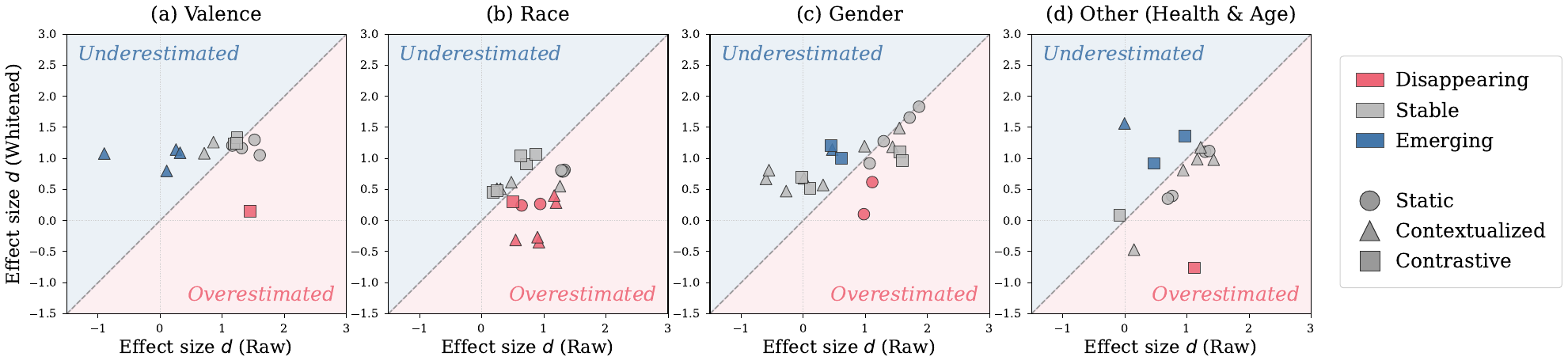}
\caption{Scatter plots of WEAT effect sizes before (x-axis) and after (y-axis) ZCA whitening, separated by bias type: (a)~Valence, (b)~Race, (c)~Gender, and (d)~Other (Health \& Age). The diagonal line represents no change; points above the line indicate increased effect sizes after whitening (underestimation in the raw space), and points below indicate decreased effect sizes (overestimation). Marker color denotes significance change: Disappearing (red), Stable (pink), and Emerging (blue). Marker shape denotes architectural family: circle (static), triangle (contextualized), and square (contrastive).}
\label{fig:weat_scatter}
\end{figure*}

Gender measurements show considerable scatter around the diagonal in Figure~\ref{fig:weat_scatter}(c), reflecting coexistence of overestimated and underestimated bias.
Static models retain strong significance on W6 (word2vec: $d = 1.89 \to 1.81$; GloVe: $d = 1.72 \to 1.63$), but word2vec shows Disappearing cases on W7 and W8. The contextualized models reveal a more nuanced pattern: only one of nine combinations shows Emerging bias (BERT on W8: $d = 0.48 \to 1.15$). GPT-2 shows directional reversals on W7 ($d = -0.58 \to 0.69$) and W8 ($d = -0.54 \to 0.81$), where raw associations flip from negative to positive after whitening. This suggests that severe anisotropy can invert the apparent sign of association.
The contrastive models provide further evidence: unsup-BERT shows Emerging bias on both W7 ($d = 0.47 \to 1.18$) and W8 ($d = 0.59 \to 0.98$).
These emerging effect sizes are comparable to or larger than those in the standard contextualized models, suggesting that while contrastive training reduces geometric distortion, it does not eliminate the bias encoded in the pre-training data.

\subsubsection{Health and Age (W9--W10).}

For health (W9), the most notable result is GPT-2's Emerging case ($d = -0.01 \to d = 1.55$), where whitening uncovers a strong association that was geometrically masked in the raw space.
Unsup-BERT also shows an Emerging case and unsup-RoBERTa a Disappearing one on W9, while static models show mixed patterns (see Appendix~C).
For age (W10), six of seven models show Stable results, with one Emerging case (unsup-RoBERTa: $d = 0.49 \to d = 0.92$); the contextualized models maintain large significant effects ($d > 1.0$ after whitening).

\subsection{The Role of Anisotropy Severity}

The frequency of significance changes appears broadly consistent with the degree of initial anisotropy: GPT-2 (mean cosine 0.997) shows changes in 5 of 10 tests, BERT (0.821) in 5, unsup-BERT (0.386) in 4, while word2vec (0.059) shows changes in 3 of 10 tests. However, RoBERTa (0.954) shows only 1 change despite its high anisotropy, and GloVe (0.192) likewise shows only 1 change, indicating that the relationship between anisotropy severity and the frequency of significance changes is not strictly monotonic and may depend on additional factors such as the specific geometry of the embedding space relative to the WEAT stimulus words.

\subsection{Patterns of Distortion Across Bias Categories and Embedding Families}

The category-level analysis reveals two systematic asymmetries, summarized visually in Figure~\ref{fig:weat_scatter}.
First, distortion direction depends on bias category: Disappearing cases
concentrate in the race tests, which account for 8 of 12 (66.7\%), while
Emerging cases are most frequent in the valence tests (4 of 10), with the
remainder split evenly between gender (3) and health/age (3).
This asymmetry suggests that anisotropy may interact differently with different stimulus configurations. However, the specific mechanism behind this pattern is not fully clear from our experimental results alone. One possibility is that the geometric relationship between target sets and the dominant variance directions of the embedding space differs across bias categories. As a result, anisotropy may amplify cosine differences in some cases and compress them in others. Identifying the precise factors that determine the direction of distortion is an important direction for future work.

Second, distortion direction depends on architectural family.
The contextualized models, which exhibit the highest anisotropy (mean cosine 0.821--0.997), account for 6 of 10 Emerging cases (60.0\%), consistent with the interpretation that severe anisotropy compresses the cosine similarity range until genuine differential associations fall below the significance threshold.
The static models, despite their relatively low anisotropy, contribute 4 of 12 Disappearing cases (33.3\%), indicating that even moderate distortion can inflate specific measurements.
The contrastive models exhibit both directions of distortion across categories, indicating that contrastive training reduces geometric distortion but neither prevents it
uniformly nor eliminates the bias encoded in the pre-training data.

%==============================================================
\section{Discussion}
%==============================================================

\subsection{Implications for Computational Social Science}
\label{implications}

Our findings have direct implications for intrinsic bias measurement.
The observation that approximately 30\% of WEAT measurements change significance status after geometric calibration suggests that a substantial fraction of assessments based on uncalibrated embeddings may be unreliable.
Practitioners conducting bias audits should therefore perform an isotropy check before applying cosine-based measurements and calibrate when anisotropy is detected.

The WEAT has been widely used in computational social science to study societal biases that are difficult or costly to measure experimentally, including historical shifts in stereotypes over 200 years \citep{charlesworth2022historical}, 100 years of gender and ethnic stereotypes \citep{garg2018word}, gender stereotypes across 25 languages \citep{lewis2020gender}, and intersectional stereotypes \citep{charlesworth2024extracting}.
Our results suggest that conclusions drawn from highly anisotropic models may need to be revisited with calibrated measurements to determine whether reported biases were overestimated, underestimated, or accurately captured.
This does not invalidate prior findings; rather, geometric calibration offers a tool to assess their robustness.

The bidirectional nature of the distortion makes this concern particularly pressing.
If anisotropy only inflated bias measurements, uncalibrated results could be interpreted as upper bounds.
However, because anisotropy can also mask genuine biases, uncalibrated results cannot be treated as conservative estimates either.

Our work parallels methodological debates in psychology about how implicit bias should be measured.
Just as \citet{greenwald2022best} provided recommended best practices for research using the IAT, we argue that computational bias measures require analogous methodological discipline, including correction for known geometric distortions.
\citet{morehouse2025rethinking} likewise called for importing social science best practices into LLM bias probing.
Our ZCA calibration addresses one specific aspect of measurement invariance: ensuring that WEAT scores are comparable across models with differing degrees of anisotropy.

The importance of calibrating intrinsic bias measurements is reinforced by growing evidence that such biases propagate to downstream behavior.
EAT-measured biases in vision-language models correlate with downstream task performance and propagate to zero-shot retrieval \citep{ghate2025biases}, image classification \citep{wolfe2022markedness}, visual question answering, captioning, and generation \citep{wolfe2022american,friedrich2024auditing}, and sentiment classification \citep{mei2023bias}.
Biased AI outputs have also been shown to shape humans' own implicit
associations \citep{sim2025biased}, establishing a pathway from
embedding-level bias to changes in human implicit associations.
If intrinsic measurements are distorted by anisotropy, the resulting inaccuracies may lead to erroneous assessments of downstream risk, making geometric calibration a prerequisite for reliable bias auditing.

As AI systems come under regulatory oversight, bias auditing tools must meet high standards of reliability.
The EU AI Act \citep{euaiact2024} mandates bias examination as part of its data governance requirements for datasets used in high-risk AI systems, and the NIST AI Risk Management Framework \citep{nist2023ai} emphasizes valid and reliable measurement in AI risk assessment.
ZCA calibration could be integrated into such frameworks as a standard preprocessing step for embedding-based bias measurement, ensuring comparable audit results across models.

\subsection{Compatibility with EAT Variants}

ZCA whitening operates as a pre-processing calibration step that is agnostic to the specific form of the association test applied afterward.
Because it transforms only the embedding space geometry without modifying the test procedure itself, it is compatible with any cosine-similarity-based bias metric.
It can thus be applied to sentence embeddings in SEAT \citep{may2019measuring}, contextualized vectors in CEAT \citep{guo2021detecting}, image embeddings in iEAT \citep{steed2021image}, and speech representations in SpEAT \citep{slaughter2023pre}.
For ML-EAT \citep{wolfe2024mleat}, our approach provides a corrective step that complements its anisotropy diagnostic: when ML-EAT identifies a model as highly anisotropic, ZCA whitening can restore the conditions under which cosine-based evaluation becomes meaningful.
More broadly, any future variant of embedding association tests relying on cosine similarity would benefit from the same calibration procedure.

\subsection{Limitations}

Several limitations should be acknowledged.
First, the whitening matrix is estimated from WikiText-103, and different reference corpora may yield different transformations; the sensitivity of WEAT outcomes to the choice and size of the reference corpus warrants systematic investigation.
Second, our evaluation is limited to seven models, and whether the calibration behavior we observe generalizes to a broader range of models requires further investigation.
Third, our implementation uses a fixed regularization parameter $\varepsilon$ for numerical stability, and we do not provide a sensitivity analysis of how this choice affects calibrated WEAT scores.
Fourth, the relationship between calibrated WEAT measurements and downstream task bias remains an open question. Prior work has reported mixed findings: \citet{goldfarb2021intrinsic}  found frequent divergence, while \citet{ghate2025biases} demonstrated correlation in vision-language models. Whether calibration improves this relationship requires future investigation.
Addressing these open questions constitutes an important direction for future research.

Beyond these limitations, we emphasize the scope of our contribution.
This work evaluates the \emph{geometric reliability} of WEAT
measurements. We examine whether cosine similarity is a valid metric
within the embedding space, not how the associations it measures shape
the outputs of a deployed system.
We do not evaluate the alignment between WEAT measurements and human implicit association measures.
The latter is a separate question that applies equally to raw and calibrated measurements and has been examined in prior work \citep{caliskan2017semantics, morehouse2023traces,
charlesworth2024echoes}.
Extending human alignment analysis to calibrated measurements is a natural next step.

%==============================================================
\section{Conclusion}
This research proposed ZCA whitening as a pre-processing calibration step for
the Word Embedding Association Test (WEAT). The method transforms the covariance
of the embedding space into the identity matrix while minimizing perturbation to
the original vectors, restoring the isotropy condition on which WEAT relies.
We evaluated our approach on ten WEAT test suites and seven models
(70 model--task combinations). ZCA whitening reduced anisotropy across all
models, and for highly anisotropic models it also improved standard semantic
similarity benchmark scores. After calibration, over 30\% of WEAT results
changed significance status, and effect sizes shifted in both directions
depending on bias category.

These results suggest that bias measurements obtained in strongly anisotropic
spaces may partly reflect geometric properties of the space rather than social
associations alone. We do not claim that previously reported findings are
invalid. Rather, studies conducted under such conditions may benefit from
re-examination with a calibrated metric.

ZCA whitening operates as a model-agnostic pre-processing step. It can be
applied not only to WEAT but also to its variants, such as CEAT, iEAT, and
SpEAT, and in principle to any bias metric that relies on cosine similarity.
The reliability of such measurements is inseparable from the geometry of the
space in which they are computed, and our approach offers a practical step
toward geometrically valid bias measurement in computational social science and
AI fairness research.

%==============================================================
%\section*{Ethical Considerations}
\section*{Ethical Statement}

This work aims to improve the reliability of bias measurement in word embeddings. However, several ethical considerations deserve attention.

\paragraph{Risk of misinterpretation.}
Our finding that some previously significant bias measurements may not survive geometric calibration could be misinterpreted as evidence that AI systems are less biased than previously reported.
We caution against this interpretation.
Calibration reveals that while some specific measurements were inflated, others were deflated, and the overall picture is one of measurement unreliability rather than bias absence.
The appropriate response is to re-measure with calibrated tools, not to conclude that bias is less prevalent.
This concern is particularly acute given the bidirectional nature of the distortion we document: practitioners, journalists, or model developers who selectively cite Disappearing cases as evidence of model improvement would be misrepresenting our findings.
As \citet{blodgett2020language} argued, NLP bias research must be grounded in clear normative reasoning about who is harmed and how; our calibration method addresses one specific source of measurement error but does not resolve deeper questions about what constitutes bias or how bias measurements should inform practice.

\paragraph{Scope of intrinsic bias metrics.}
Even with proper geometric calibration, intrinsic bias measurements capture only one dimension of the complex ways in which AI systems can perpetuate social inequities.
Embedding-level bias metrics should be used alongside downstream task evaluations \citep{goldfarb2021intrinsic,cabello2023independence}, qualitative audits, and participatory assessments involving affected communities \citep{selbst2019fairness}.
The communities most likely to be harmed by biased AI systems include racial, gender, and other minoritized groups represented in WEAT target sets. Their perspectives should inform what counts as a meaningful bias measurement in the first place.

\paragraph{Responsibility in auditing contexts.}
As discussed in Section~\ref{implications}, embedding-based bias measurements are increasingly used in regulatory and auditing contexts.
This dual-use character introduces specific responsibilities: a calibration method that changes which biases are flagged as significant could, if applied uncritically, either expose previously hidden harms or obscure documented ones.
We therefore recommend that calibrated measurements be reported alongside uncalibrated ones rather than replacing them, so that audit trails preserve both perspectives and allow stakeholders to assess the geometric reliability of each measurement.

\section*{Acknowledgments}

We thank the reviewers of AIES 2026 for their thoughtful and
constructive feedback. We also thank our colleagues for helpful discussions throughout this work.

\bibliography{references}

\clearpage
\setcounter{section}{0}
\renewcommand{\thesection}{\Alph{section}}
\setcounter{figure}{0}\renewcommand{\thefigure}{A\arabic{figure}}
\setcounter{table}{0}\renewcommand{\thetable}{A\arabic{table}}
\setcounter{equation}{0}\renewcommand{\theequation}{A\arabic{equation}}

\twocolumn[
  \begin{center}
    {\Large\bfseries Supplementary Material for:\\[0.3em]
     Calibrating WEAT Against Anisotropy: ZCA Whitening as a\\
     Geometric Pre-Processing Step for Embedding Association Tests\par}
    \vspace{1.5em}
  \end{center}
]

\setcounter{figure}{0}\renewcommand{\thefigure}{A\arabic{figure}}
\setcounter{table}{0}\renewcommand{\thetable}{A\arabic{table}}
\setcounter{equation}{0}\renewcommand{\theequation}{A\arabic{equation}}

\appendix

\section{Derivation of the Identity Covariance Property of ZCA-Whitened Vectors}
\label{app:zca_proof}

This appendix provides the complete derivation showing that the ZCA-whitened vectors $z_i = W_{\text{ZCA}}(x_i - \mu)$ have identity covariance, as summarized in Equation~10 of the main paper.

Starting from the definition of the sample covariance of $\{z_i\}$:
\begin{align}
\operatorname{Cov}(z) &= \frac{1}{n-1}\sum_{i=1}^n z_i z_i^\top \notag \\
&= \frac{1}{n-1}\sum_{i=1}^n W_{\text{ZCA}}(x_i - \mu)\bigl(W_{\text{ZCA}}(x_i - \mu)\bigr)^\top \notag \\
&= \frac{1}{n-1}\sum_{i=1}^n W_{\text{ZCA}}(x_i - \mu)(x_i - \mu)^\top W_{\text{ZCA}}^\top \notag \\
&= W_{\text{ZCA}}\left(\frac{1}{n-1}\sum_{i=1}^n (x_i - \mu)(x_i - \mu)^\top\right) W_{\text{ZCA}}^\top \notag \\
&= W_{\text{ZCA}}\,\Sigma\, W_{\text{ZCA}}^\top \notag \\
&= \Sigma^{-1/2}\, \Sigma\, \Sigma^{-1/2} \notag \\
&= I.
\label{eq:identity_proof_appendix}
\end{align}
The third equality uses the transpose identity $(AB)^\top = B^\top A^\top$.
The sixth equality substitutes $W_{\text{ZCA}} = \Sigma^{-1/2}$, using the fact that this matrix is symmetric (so $W_{\text{ZCA}}^\top = W_{\text{ZCA}}$).
The final equality follows from
\begin{equation*}
\Sigma^{-1/2}\, \Sigma\, \Sigma^{-1/2} = \Sigma^{-1/2}\, \Sigma^{1/2}\, \Sigma^{1/2}\, \Sigma^{-1/2} = I.
\end{equation*}

\newpage

\section{Pairwise Cosine Similarity Heatmaps for All Models}
\label{app:cosine_heatmaps}

\begin{figure}[htbp]
\centering
\includegraphics[width=\columnwidth]{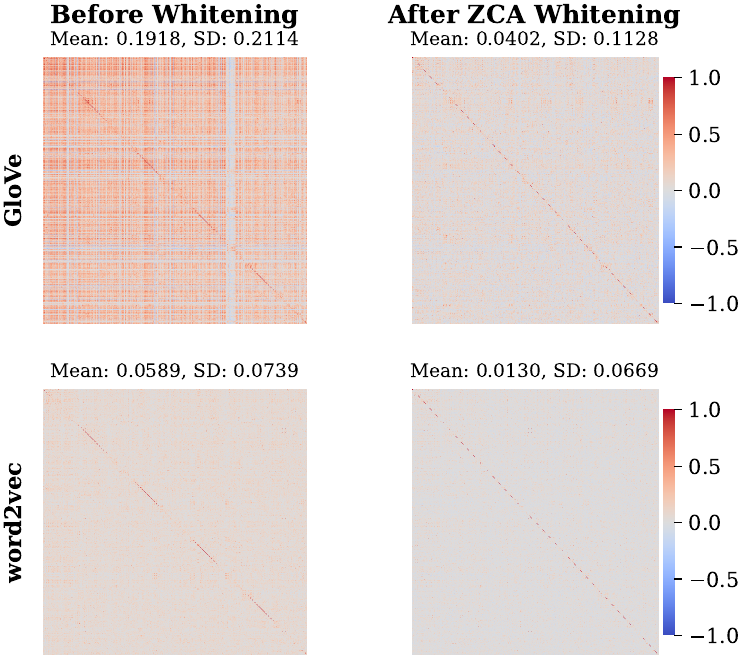}
\caption{Pairwise cosine similarity matrices for static embedding models (GloVe, word2vec) before (left) and after (right) ZCA whitening.}
\label{fig:cosine_static}
\end{figure}

\begin{figure}[htbp]
\centering
\includegraphics[width=\columnwidth]{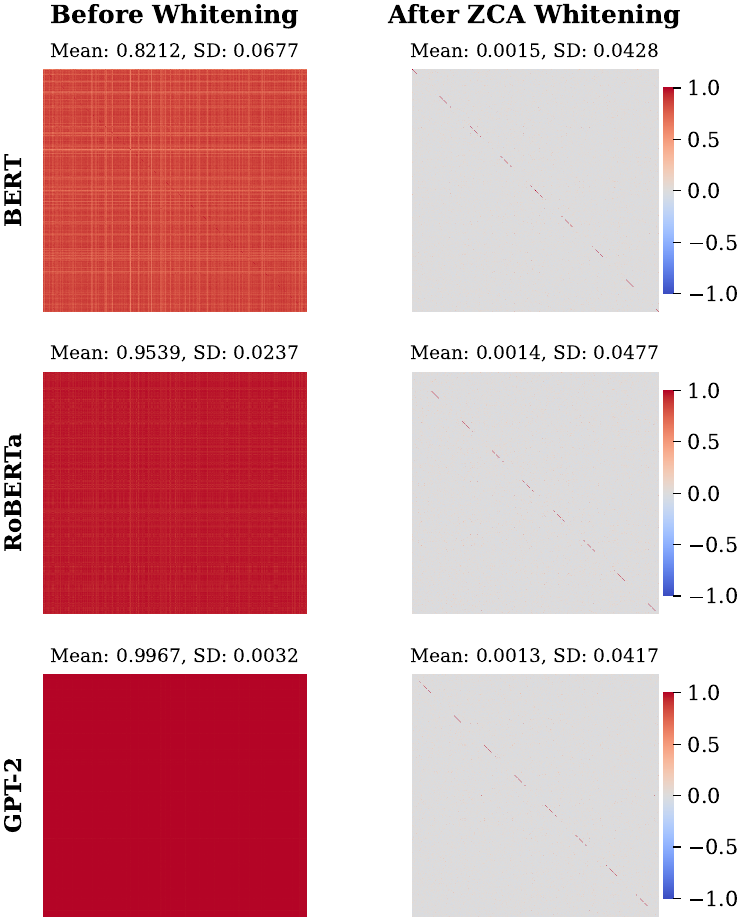}
\caption{Pairwise cosine similarity matrices for contextualized models (BERT, RoBERTa, GPT-2) before (left) and after (right) ZCA whitening.}
\label{fig:cosine_contextualized}
\end{figure}

\begin{figure}[htbp]
\centering
\includegraphics[width=\columnwidth]{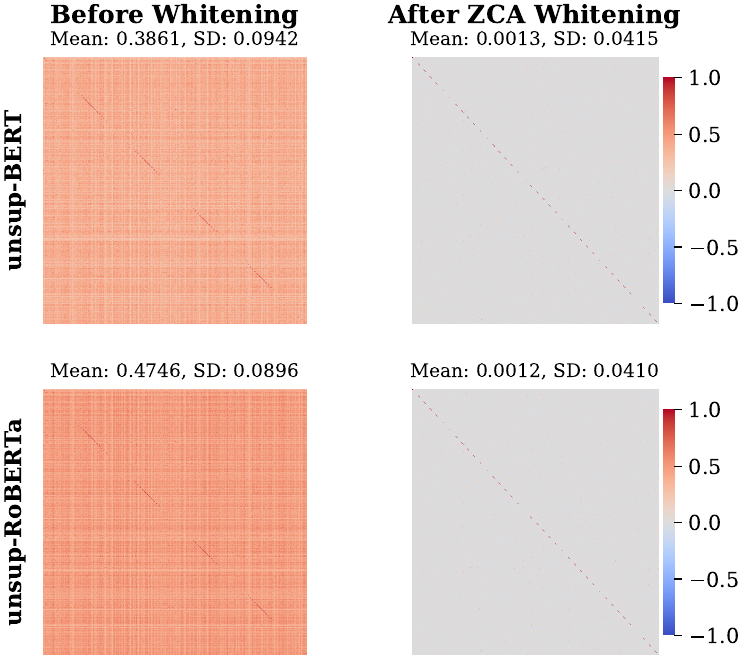}
\caption{Pairwise cosine similarity matrices for contrastive models (unsup-BERT, unsup-RoBERTa) before (left) and after (right) ZCA whitening.}
\label{fig:cosine_contrastive}
\end{figure}

Figures~\ref{fig:cosine_static}--\ref{fig:cosine_contrastive} show pairwise cosine similarity matrices for 2{,}000 randomly sampled embedding vectors, grouped by architectural family. Across all seven models, ZCA whitening substantially reduces pairwise similarity regardless of the initial anisotropy severity.

\clearpage
\onecolumn

\section{Full WEAT Results}
\label{app:weat_results}
% ============================================================
% TABLE 4: WEAT Results
% ============================================================
%\begin{table}[H]
\begingroup
\footnotesize
\setlength{\tabcolsep}{4pt}

\centering
\captionof{table}{WEAT effect sizes ($d$) and permutation $p$-values (100,000 permutations, one-sided) before (Raw) and after (White) ZCA whitening across seven models and ten test suites (see Table~1 for full definitions). Significance at $p < 0.05$. \textbf{S}~=~Stable, \textbf{D}~=~Disappearing (significant $\to$ non-significant), \textbf{E}~=~Emerging (non-significant $\to$ significant). Bold $d$ values indicate $p < 0.05$. $p$-values smaller than $10^{-4}$ are reported as their order of magnitude.}
\label{tab:weat_results}
\footnotesize
\setlength{\tabcolsep}{4pt}

\vspace{0.3em}
\textbf{Static Embeddings}
\vspace{0.2em}

\begin{tabular}{@{}llrrrrc rrrrc@{}}
\toprule
& & \multicolumn{5}{c}{\textbf{GloVe}} & \multicolumn{5}{c}{\textbf{word2vec}} \\
\cmidrule(lr){3-7} \cmidrule(lr){8-12}
\textbf{ID} & \textbf{Bias} & $d_R$ & $p_R$ & $d_W$ & $p_W$ & Chg & $d_R$ & $p_R$ & $d_W$ & $p_W$ & Chg \\
\midrule
W1  & Valence & \textbf{1.17} & $10^{-5}$ & \textbf{1.18} & $10^{-5}$ & S & \textbf{1.54} & $10^{-5}$ & \textbf{1.32} & $10^{-5}$ & S \\
W2  & Valence & \textbf{1.30} & $10^{-5}$ & \textbf{1.16} & $10^{-5}$ & S & \textbf{1.63} & $10^{-5}$ & \textbf{1.03} & $10^{-5}$ & S \\
W3  & Race    & \textbf{0.95} & $10^{-5}$ & 0.24 & .1660 & D & \textbf{0.67} & .0031 & 0.22 & .1989 & D \\
W4  & Race    & \textbf{1.32} & $10^{-5}$ & \textbf{0.81} & .0098 & S & \textbf{1.31} & $10^{-5}$ & \textbf{0.78} & .0125 & S \\
W5  & Race    & \textbf{1.32} & $10^{-5}$ & \textbf{0.81} & .0095 & S & \textbf{1.31} & $10^{-5}$ & \textbf{0.78} & .0131 & S \\
W6  & Gender  & \textbf{1.72} & $10^{-5}$ & \textbf{1.63} & $10^{-5}$ & S & \textbf{1.89} & $10^{-5}$ & \textbf{1.81} & $10^{-5}$ & S \\
W7  & Gender  & \textbf{1.28} & .0020 & \textbf{1.27} & .0023 & S & \textbf{0.97} & .0233 & 0.09 & .4353 & D \\
W8  & Gender  & \textbf{1.09} & .0091 & \textbf{0.94} & .0280 & S & \textbf{1.14} & .0100 & 0.59 & .1263 & D \\
W9  & Health  & \textbf{1.30} & .0069 & \textbf{1.08} & .0254 & S & \textbf{1.38} & .0025 & \textbf{1.14} & .0148 & S \\
W10 & Age     & 0.75 & .0691 & 0.39 & .2294 & S & 0.72 & .0769 & 0.33 & .2636 & S \\
\bottomrule
\end{tabular}

\vspace{0.6em}
\textbf{Contextualized Embeddings}
\vspace{0.2em}

\begin{tabular}{@{}llrrrrc rrrrc rrrrc@{}}
\toprule
& & \multicolumn{5}{c}{\textbf{GPT-2}} & \multicolumn{5}{c}{\textbf{BERT}} & \multicolumn{5}{c}{\textbf{RoBERTa}} \\
\cmidrule(lr){3-7} \cmidrule(lr){8-12} \cmidrule(lr){13-17}
\textbf{ID} & \textbf{Bias} & $d_R$ & $p_R$ & $d_W$ & $p_W$ & Chg & $d_R$ & $p_R$ & $d_W$ & $p_W$ & Chg & $d_R$ & $p_R$ & $d_W$ & $p_W$ & Chg \\
\midrule
W1  & Valence & 0.25 & .1869 & \textbf{1.13} & $10^{-5}$ & E & \textbf{0.69} & .0063 & \textbf{1.10} & $10^{-5}$ & S & \textbf{0.87} & .0008 & \textbf{1.26} & $10^{-5}$ & S \\
W2  & Valence & $-$0.87 & .9993 & \textbf{1.05} & $10^{-5}$ & E & 0.34 & .1211 & \textbf{1.11} & $10^{-5}$ & E & 0.11 & .3486 & \textbf{0.81} & .0018 & E \\
W3  & Race    & \textbf{1.24} & $10^{-5}$ & \textbf{0.57} & .0109 & S & \textbf{0.55} & .0142 & $-$0.30 & .8829 & D & \textbf{0.48} & .0272 & \textbf{0.60} & .0082 & S \\
W4  & Race    & \textbf{0.94} & .0022 & $-$0.33 & .8209 & D & \textbf{1.19} & $10^{-5}$ & 0.31 & .1886 & D & 0.32 & .1867 & 0.51 & .0772 & S \\
W5  & Race    & \textbf{0.91} & .0032 & $-$0.27 & .7706 & D & \textbf{1.18} & .0001 & 0.41 & .1256 & D & 0.24 & .2530 & 0.54 & .0644 & S \\
W6  & Gender  & 0.30 & .2901 & 0.59 & .1259 & S & \textbf{0.99} & .0217 & \textbf{1.21} & .0051 & S & \textbf{1.44} & .0008 & \textbf{1.17} & .0077 & S \\
W7  & Gender  & $-$0.58 & .8641 & 0.69 & .0867 & S & \textbf{1.54} & .0003 & \textbf{1.51} & .0007 & S & 0.03 & .4803 & 0.68 & .0880 & S \\
W8  & Gender  & $-$0.54 & .8531 & 0.81 & .0525 & S & 0.48 & .1742 & \textbf{1.15} & .0084 & E & $-$0.28 & .7113 & 0.50 & .1658 & S \\
W9  & Health  & $-$0.01 & .5044 & \textbf{1.55} & .0011 & E & 0.92 & .0573 & 0.83 & .0800 & S & 0.16 & .3974 & $-$0.47 & .7795 & S \\
W10 & Age     & \textbf{1.25} & .0057 & \textbf{1.15} & .0058 & S & \textbf{1.45} & .0004 & \textbf{1.00} & .0208 & S & \textbf{1.17} & .0060 & \textbf{1.00} & .0225 & S \\
\bottomrule
\end{tabular}

\vspace{0.6em}
\textbf{Unsupervised Contrastive Embeddings}
\vspace{0.2em}

\begin{tabular}{@{}llrrrrc rrrrc@{}}
\toprule
& & \multicolumn{5}{c}{\textbf{unsup-BERT}} & \multicolumn{5}{c}{\textbf{unsup-RoBERTa}} \\
\cmidrule(lr){3-7} \cmidrule(lr){8-12}
\textbf{ID} & \textbf{Bias} & $d_R$ & $p_R$ & $d_W$ & $p_W$ & Chg & $d_R$ & $p_R$ & $d_W$ & $p_W$ & Chg \\
\midrule
W1  & Valence & \textbf{1.18} & .0008 & \textbf{1.26} & $10^{-5}$ & S & \textbf{1.23} & $10^{-5}$ & \textbf{1.23} & $10^{-5}$ & S \\
W2  & Valence & \textbf{1.25} & $10^{-5}$ & \textbf{1.34} & $10^{-5}$ & S & \textbf{1.47} & $10^{-5}$ & 0.15 & .2990 & D \\
W3  & Race    & \textbf{0.50} & .0220 & 0.32 & .1002 & D & \textbf{0.73} & .0014 & \textbf{0.93} & $10^{-5}$ & S \\
W4  & Race    & 0.21 & .2792 & 0.43 & .1132 & S & \textbf{0.86} & .0059 & \textbf{1.07} & .0007 & S \\
W5  & Race    & 0.22 & .2725 & 0.46 & .1006 & S & \textbf{0.65} & .0315 & \textbf{1.04} & .0009 & S \\
W6  & Gender  & \textbf{1.55} & .0005 & \textbf{1.12} & .0099 & S & \textbf{1.61} & $10^{-4}$ & \textbf{0.99} & .0218 & S \\
W7  & Gender  & 0.47 & .1887 & \textbf{1.18} & .0055 & E & 0.10 & .4236 & 0.52 & .1553 & S \\
W8  & Gender  & 0.59 & .1277 & \textbf{0.98} & .0185 & E & $-$0.00 & .5038 & 0.70 & .0864 & S \\
W9  & Health  & 0.96 & .0517 & \textbf{1.38} & .0020 & E & \textbf{1.12} & .0228 & $-$0.78 & .9073 & D \\
W10 & Age     & $-$0.07 & .5591 & 0.09 & .4343 & S & 0.49 & .1823 & 0.92 & .0314 & E \\
\bottomrule
\end{tabular}
%\end{table}

\endgroup

\end{document}